\documentclass[conference]{IEEEtran}
\IEEEoverridecommandlockouts
\usepackage[T1]{fontenc}
\usepackage{cite}
\usepackage{amsmath,amssymb,amsfonts}
\usepackage{algorithm}
\usepackage{algpseudocode}   
\usepackage{booktabs}
\usepackage{stfloats}

\algrenewcommand\algorithmicrequire{\textbf{Input:}}
\algrenewcommand\algorithmicensure{\textbf{Output:}}

\usepackage{graphicx}
\usepackage{subcaption}
\usepackage{textcomp}
\usepackage{xcolor}
\usepackage{xurl}   
\usepackage[hidelinks]{hyperref}
\def\BibTeX{{\rm B\kern-.05em{\sc i\kern-.025em b}\kern-.08em
    T\kern-.1667em\lower.7ex\hbox{E}\kern-.125emX}}

\usepackage{tikz}

\newcommand{\circled}[1]{%
  \tikz[baseline=(c.base)]\node[draw,circle,fill=black,inner sep=1pt] (c)
  {\textcolor{white}{\bfseries #1}};%
}

\begin{document}

\title{DBLP: Phase-Aware Bounded-Loss Transport for Burst-Resilient Distributed ML Training \\
}

\newif\ifanonymous

\ifanonymous
  \author{Anonymous Authors}
\else
    \author{
        \IEEEauthorblockN{
            Zechen Ma\textsuperscript{*},
            Zixi Qu\textsuperscript{*},
            Jinyan Yi\textsuperscript{*},
            David Lin,
            Yashar Ganjali
        }
        \IEEEauthorblockA{
            Department of Computer Science, University of Toronto, Toronto, Canada\\
            \{zma, zixiqu, jyi, davidlin, yganjali\}@cs.toronto.edu\\
            \textsuperscript{*}Equal contribution
        }
    }
\fi

\maketitle 

\begin{abstract}

Distributed machine learning (ML) training has become a necessity with the prevalence of billion to trillion-parameter-scale models. As these workloads scale out across larger clusters, and increasingly across sites, classical issues such as bursts and long-tail latency are amplified across the interconnect stack. While prior work has improved training efficiency from the ML perspective at the application layer, it often fails to address transient congestion events at the network layer that introduce severe long-tail latency and training-time variability. Existing network optimizations treat all gradients equally and thus fail to integrate sufficient model-training insights into communication protocol design.

In this paper, we present \textbf{Dynamic Bounded-Loss Protocol (DBLP)}, a burst-resilient, training-phase-aware, and hardware-agnostic transport-layer intervention that incorporates model-level tolerance properties into gradient communication. By dynamically adjusting gradient loss tolerance across training phases, DBLP reduces overall training time and mitigates tail latency collapse during transient high-loss events. 

Compared to the current state-of-the-art solution, DBLP tolerates significantly higher loss while achieving comparable test accuracy, and reduces end-to-end training time by an average of $24.8\%$ and a maximum of $33.9\%$. At microburst events, DBLP achieves up to $5.97\times$ single-round communication latency speedups over the baseline, preventing burst-induced long-tail latency spikes and maintaining stable training performance.

\end{abstract}


\section{Introduction}

Recent advancements in deep learning have been accompanied by rapid growth in model size. Neural networks with billions to trillions of parameters have emerged in natural language processing~\cite{openai2024gpt4technicalreport} and computer vision~\cite{lu2024deepseekvlrealworldvisionlanguageunderstanding} tasks. These models consistently demonstrate that scaling up parameters improves performance, but the resulting scale necessitates distributed training.

The overhead of gradient communication poses a key scalability challenge in distributed training of large-scale models~\cite{pipedream, sigcomm21}. For instance, COMET~\cite{comet} found that communication accounted for $47\%$ of the execution time in the forward pass. To mitigate this, prior work has explored techniques such as overlapping communication with computation to partially hide latency~\cite{Jayarajan2019PrioritybasedPP}, as well as gradient compression~\cite{deepgradientcompression}, quantization~\cite{quantization}, sparsification~\cite{sparsification}, and low-rank gradient compression~\cite{lowrank}. While these solutions effectively reduce the average latency, the tail latency (often caused by packet loss, network congestion, and transient traffic bursts) remains largely unaddressed~\cite{mlt}. Such tail behavior introduces significant training-time variability and undermines the quality of service of distributed deep neural network (DNN) training~\cite{tail-qos}.

Based on these observations, we identify five design properties that a modern gradient communication protocol should satisfy: \circled{1} Order-independence: the protocol should not rely on in-order delivery; gradients should be applied as soon as they arrive. \circled{2} Bounded-loss tolerance: the system should leverage the model’s intrinsic tolerance to bounded gradient loss, thereby avoiding excessive retransmissions that contribute to long-tail latency. \circled{3} Burst-resilience: since stochastic and short-lived microbursts are inevitable in data centers~\cite{micro-load-balancing, short-lived-1, short-lived-2}, the protocol should maintain robust performance under such events. \circled{4} Phase adaptivity: as a model’s sensitivity to gradient loss varies across training phases, the communication layer should dynamically adapt its tolerance level over time. \circled{5} Hardware agnosticism: the protocol should minimize reliance on configurable network infrastructure, enabling deployment across commodity clusters and cloud environments where switch-level configuration is infeasible.

Prior work, Machine Learning Transport (MLT)~\cite{mlt}, proposed a customized network transport protocol that optimizes both average and tail latency in distributed DNN training. They leverage inter-packet dependency and priority-based packet queuing and dropping on switches to achieve properties \circled{1} and \circled{2}, which violates~\circled{5}. Property \circled{3} is claimed to be satisfied; however, our microburst experiments show that an implementation of MLT without hardware support fails to sustain its communication latency and goodput during burst events. Furthermore, \circled{4} remains unaddressed. 

Phase adaptivity \circled{4} is supported by prior work: early training phases are sensitive to corrupted inputs, leading to irreversible performance degradation~\cite{achille2019criticallearningperiodsdeep}. Building on this insight, later work~\cite{Accordion} introduces gradient-norm-based critical-phase detection and applies it to gradient compression and batch-size scheduling. These results suggest that communication reliability decisions should be informed by training dynamics, motivating tighter co-design between the algorithm layer and the transport stack.

To satisfy all five properties, we present \textbf{Dynamic Bounded-Loss Protocol (DBLP)}, a transport-layer intervention for distributed gradient communication. DBLP natively supports out-of-order packet delivery over unreliable channels. It exploits model-level loss tolerance by transmitting only a subset of gradient packets within a controlled threshold. DBLP further leverages gradient-norm fluctuations to dynamically adjust this threshold across training phases, in response to the model's varying sensitivity to gradient loss. Unlike prior work~\cite{mlt} that relies on specialized switch configurations, DBLP requires no hardware modifications, making it deployable across the commodity fabrics that connect modern scale-up, scale-out, and scale-across training deployments.

Our experiments demonstrate that DBLP preserves model accuracy while improving training efficiency and stability. When trained for the same number of epochs, DBLP tolerates $40\%$ additional gradient loss compared to the baseline while achieving comparable evaluation accuracy. Across all evaluated models, DBLP reduces end-to-end training time by an average of $24.8\%$ and a maximum of $33.9\%$ over the baseline. Under microburst events, DBLP achieves up to $5.97\times$ latency speedups within a single round of bursty communication, delivering up to $4.17\times$ lower tail latency and $1.95\times$ lower average latency than the baseline. These results show that aligning transport decisions with training dynamics effectively mitigates long-tail latency issues amplified as training scales out across larger clusters and sites.  

\section{Background and Motivation}
\label{background-and-motivation}

\subsection{Communication As a Bottleneck}
\subsubsection{Distributed DNN Training Basics}

DNNs train iteratively: each mini-batch is forward-propagated to compute a loss, gradients are derived via backpropagation, and model parameters are updated.

As models have scaled to billions or trillions of parameters~\cite{openai2024gpt4technicalreport}, single-accelerator training is no longer feasible, necessitating distributed parallelism strategies. Prior work has explored a range of approaches, including data parallelism~\cite{dp}, tensor parallelism~\cite{megatron-lm}, pipeline parallelism~\cite{pipedream, gpipe}, and expert parallelism~\cite{gshard}. As our focus is on DBLP's protocol behavior, we adopt data parallelism and leave other schemes to future work. 

Data parallelism partitions the training data across nodes with each node maintaining a complete replica of the model. In data-parallel training, each worker node computes gradients locally and must periodically synchronize them across nodes to ensure consistent model updates. This synchronization can be performed through collective or centralized routines. Recent work~\cite{mixnet} pointed out that MoE training involves both decentralized (i.e., inter-host and intra-host all-to-all) and centralized (i.e., intra-host gather and scatter) communication patterns. As a result, we built a centralized all-reduce architecture that supports reduce and broadcast operations between workers and the server.

\subsubsection{Communication Overhead}

Communication between nodes demands substantial network support, as gradient exchanges per iteration can range from MB to GB~\cite{bytescheduler}, a pressure that only intensifies as jobs scale across the interconnect stack. Many prior studies have reported substantial communication overhead~\cite{comm-atc, comm3, Jayarajan2019PrioritybasedPP, bytescheduler, comm2}. For instance, Pipedream observed that training over 32 GPUs resulted in $90\%$ of total training time being spent on communication~\cite{pipedream}. Similarly, ByteScheduler found that the end-to-end performance improvement does not scale linearly with the number of GPUs due to communication bottlenecks~\cite{bytescheduler}. COMET further reinforced this observation: $47\%$ of their training time was consumed by communication~\cite{comet}.

\subsection{Existing Solutions}

MLT~\cite{mlt} identified that existing solutions for reducing the communication overhead, such as gradient compression~\cite{deepgradientcompression}, quantization~\cite{quantization}, sparsification~\cite{sparsification}, low-rank gradient compression~\cite{lowrank}, only reduced the average latency. However, the more critical contributor to training time is tail latency~\cite{DeTail}. To address this, MLT proposed bounded-loss tolerance: persistently losing a fraction of gradients does not significantly affect convergence. With tolerance $p \in [0, 1)$, training on $1-p$ of gradients achieves the same target accuracy in the same number of epochs.

With these observations in mind, we now ask the following questions: \textit{if a model has a degree of tolerance to gradient loss, will the loss be a fixed constant throughout training as suggested by MLT? Could models have varying degrees of tolerance to gradient loss across training phases?}

\subsection{Preliminary Experiment Results}

To answer these questions, we trained DenseNet169~\cite{densenet169} on CIFAR-10~\cite{cifar10} for nine epochs. We divided training into three equal phases of three epochs each: early, middle, and final. We simulated packet drops by randomly zeroing out a fraction of received gradients, repeating the experiment five times. The first run had no drops. The second dropped $40\%$ of gradients during the middle and final phases; the third repeated the second with $80\%$ drops. The last two runs applied $40\%$ and $80\%$ drops only during the early phase. The rates $40\%$ and $80\%$ were chosen arbitrarily to represent moderate and aggressive loss scenarios, respectively.

\begin{figure}[t]
    \centering
    \includegraphics[width=1.0\linewidth]{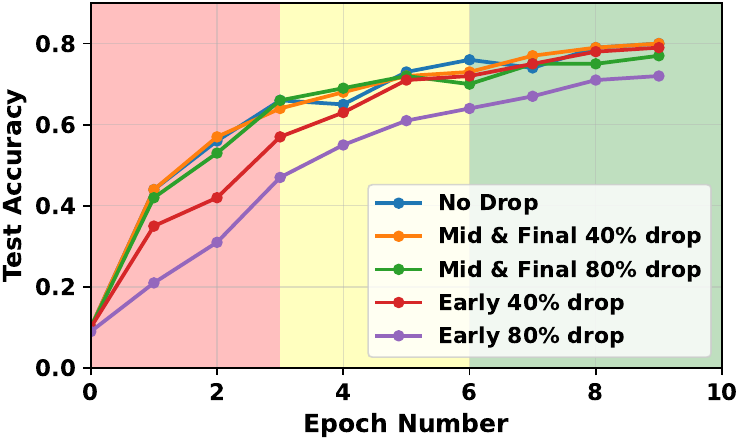}
    \caption{Test accuracy of DenseNet-169 on CIFAR-10 over nine training epochs under five gradient-drop schedules. The red, yellow, and green shaded bands denote the early, middle, and final training phases, respectively. Dropping 40\% or 80\% of gradients during the middle and final phases tracks the no-drop closely, whereas dropping the same fractions during the early phase induces a persistent accuracy gap that does not close by the end of training, motivating phase-aware loss tolerance.}
    \label{fig:preliminary-densenet}
\end{figure}

The result is shown in Figure~\ref{fig:preliminary-densenet}. When we drop $40\%$ to $80\%$ of gradients at the middle and final training phases, the model converges at roughly the same rate as when there are no drops. However, dropping gradients at the early stage significantly slows down the convergence rate, resulting in a noticeable accuracy gap.

Findings from~\cite{achille2019criticallearningperiodsdeep} are consistent with our preliminary results. Their experiments revealed that when a neural network was trained on corrupted images during early epochs, the resulting performance loss could not be recovered even with subsequent clean training. The work in~\cite{gurari2018} also demonstrates that gradients converge to a small subspace in the early stages of training. Authors from~\cite{frankle2020stabilizinglotterytickethypothesis} pointed out that sparse and trainable sub-networks emerge during the early stages as well. These findings prompt us to purposefully ask: \textit{is early phase the only stage that deserves the highest priority with the least gradient dropping tolerance? Could some sporadic iterations in other phases require a low tolerance to gradient loss as well?}

\subsection{Critical Learning Regime (CLR)}

The beginning of training is not the only critical phase. Certain conditions can trigger the model to enter a critical period that necessitates low tolerance to gradient loss.

The concept of a critical phase, introduced in~\cite{jastrzębski2018on}, can be detected at any moment using the top eigenvalues of the Hessian. The work in~\cite{Accordion} further proposes a simpler gradient-norm-based detection that is computationally cheap and yields comparable results. They scheduled the gradient compression between a high ratio $l_{high}$ and a low ratio $l_{low}$. We employ these insights in designing our transport-layer protocol, letting the training signal at the algorithm layer drive reliability decisions inside the transport stack.

\section{Design of DBLP}

In this section, we present core components of DBLP: an enhanced protocol that supports bounded-loss-tolerant transmission (\ref{DBLP as a Better BLP}); a mechanism for dynamically adapting the loss tolerance threshold (\ref{Dynamically Self-Adjusting the Loss Tolerance Value}).

\subsection{Bounded-Loss-Tolerant Transmission}
\label{DBLP as a Better BLP}

Reliable transmissions such as TCP guarantee in-order packet delivery. When packet drops occur, TCP keeps retransmitting packets until all have arrived, which can cause long-tail latency. The in-order nature also contradicts property~\circled{1}: packets should be allowed to arrive out of order. On the other hand, unreliable best-effort services such as UDP provide no delivery guarantees.


As a result, semi-reliable transmission was proposed in~\cite{mlt}. Important metadata are exchanged before training begins through reliable channels (e.g., TCP). Gradient payloads are divided into chunks and sent through unreliable channels (e.g., UDP). During a round of communication, the sender and receiver synchronize to check progress using three control signals: \textit{probe}, \textit{bitmap}, and \textit{stop}. After all gradient packets are sent, the sender transmits a probe signal. The receiver replies with a bitmap indicating delivered and missing packets. Once received packets reach the bounded threshold, the receiver sends a stop signal, ending the communication cycle. Control signals (and bitmap data) are guaranteed to be delivered through reliable channels.

UDP inherently allows out-of-order packet arrival. Prior work~\cite{mlt} divides gradients into groups and tags sub-gradients with individual IDs and offsets. This way, upon arrival, the gradients can be rearranged and the original tensor can be reconstructed. However, this approach breaks down when a packet experiences significant delay during network bursts~\cite{long-tail}. In our experiments, we indeed encountered a scenario where a packet from round $k$ arrived at the end host during round $k + 1$. There is no mechanism to distinguish packets belonging to different communication cycles. 

Therefore, we add a round counter to the UDP payload. The end host uses it to distinguish packets from different communication cycles and drops any packet that does not belong to the current round. 

\subsection{Dynamically Changing Loss Tolerance}
\label{Dynamically Self-Adjusting the Loss Tolerance Value}

When packet drops occur, both senders and receivers synchronize through bitmaps. Both prior work~\cite{mlt} and DBLP retransmit lost packets based on the most up-to-date bitmap received.  Retransmission stops once the ratio of missing packets is equal to or less than the loss value $p$.

Prior work~\cite{mlt} maintains a fixed $p$ throughout training, noting that different models have different tolerances: EfficientNetB0~\cite{efficientnet} tolerates $0.8\%$ loss without accuracy or iteration-count penalties, while ResNet50~\cite{resnet50} tolerates up to $2.4\%$. 

Inspired by our preliminary result in Figure~\ref{fig:preliminary-densenet}, we implement DBLP with an adaptive bounded-loss tolerance. Instead of dividing the training equally into three regions (early, middle, final), DBLP only has critical regions and non-critical regions. We use relative drop in gradient magnitude to detect the activation of critical regimes (the terms \textit{regime} and \textit{region} are used interchangeably throughout this paper): 

\begin{equation}
\frac{\left|\left\lVert G_{prev}\right\rVert - \left\lVert G_{curr}\right\rVert\right|}{\left\lVert G_{prev}\right\rVert} \ge \eta
\end{equation}

where $G_{prev}$ is the gradient from the previous training iteration, and $G_{curr}$ is the gradient from the current iteration. We set $ \eta = 0.5 $ following prior work~\cite{Accordion}. As~\cite{mlt} notes, larger gradients are generally more important, so we use the L2-norm, which is dominated by large gradients and attenuates small ones, making it an ideal signal for critical learning regime detection.

Once a critical phase is detected, DBLP enters a critical region with a lower tolerance $P_{low}$. We set $P_{low}$ to be the same as $p$ in~\cite{mlt}. Outside of the critical region, DBLP automatically adapts to a higher tolerance $P_{high}$.

\begin{table*}[b]
\begin{center}
\begin{small}
\begin{sc}
\resizebox{\textwidth}{!}{
\begin{tabular}{cccccc}
\toprule
\textbf{Model} & \textbf{Dataset} & \textbf{Parameters (M)} & \textbf{Baseline $p$ (\%)} & \textbf{DBLP $P_{low}$ (\%)} & \textbf{DBLP $P_{high}$ (\%)} \\
\midrule
EfficientNetB0 & CIFAR-10  & 5   & 0.8 & 0.8 & 40.8 \\
ResNet50       & CIFAR-100 & 25  & 2.4 & 2.4 & 42.4 \\
AlexNet        & CIFAR-100 & 61  & 0.8 & 0.8 & 40.8 \\
GPT-2-S & WikiText-2 & 125 & 0.8 & 0.8 & 40.8 \\
\bottomrule
\end{tabular}
}
\end{sc}
\end{small}
\end{center}
\vskip -0.1in
\caption{Per-model gradient-loss tolerance settings for the baseline and DBLP. The baseline uses a single fixed tolerance $p$ throughout training, taken from~\cite{mlt}, while DBLP applies a low tolerance $P_{low} = p$ inside the critical learning regime (CLR) and a higher tolerance $P_{high} = p + 40\%$ outside it. Parameter counts (M = million) are listed for context.}
\label{table:mlt-dblp-tolerance}
\end{table*}

\section{Implementation}
\label{Implementation}

We implement DBLP in Python and use PyTorch for model training. As DBLP does not rely on any hardware configurations, it offers a suite of software APIs that researchers can integrate into their own frameworks and deploy it on the commodity fabrics that connect modern training clusters. The APIs and implementation details are presented below.

\subsection{Metadata}
\label{impl: metadata}

Before training, DBLP synchronizes all nodes to exchange two constant values: (1) the total number of chunks per communication cycle and (2) the tensor layer names. A single exchange over a reliable channel suffices.

During training, each UDP packet includes a custom header with three fields: (1) a sequence number that uniquely identifies the packet within a single communication round and allows the end host to rearrange out-of-order packets; (2) a round counter to distinguish packets across communication rounds; and (3) the data length.

\subsection{DBLP Send}

At each training step, DBLP partitions the tensor into gradient chunks. The size of a chunk and the length of the customized metadata exactly fill up a UDP packet payload. Together with the UDP header, they form an MTU-sized packet. The total number of packets sent within one step depends on the size of the model. Once all gradients have been sent, DBLP transmits a probe signal $P$ to the receiver via TCP. The receiver responds with a signal $B$ accompanied by bitmap data, which the sender uses to retransmit only the lost packets. This process repeats until the sender receives a stop signal $S$ from the receiver. The sender monitors the reliable channel and terminates immediately upon receiving a stop signal. When a stop signal arrives before a sender finishes sending, we call it an early stop case. The pseudocode is shown in Algorithm~\ref{alg:dblp_send} in the Appendix.

\subsection{DBLP Receive}

At each training step, DBLP initializes an empty bitmap and sets the stop signal to \textit{false}. Upon receiving a valid chunk, DBLP updates the bitmap accordingly. When a probe signal $P$ arrives, the receiver replies with the most up-to-date bitmap. Once the tolerance threshold is reached, the stop signal is sent to the sender. The pseudocode is presented in Algorithm~\ref{alg:dblp_recv} in the Appendix.

\subsection{Centralized All-Reduce Architecture}

DBLP achieves the reduce and broadcast operations via a multithreaded server and three-worker setup. The goal of our evaluation is to validate the phase-aware transport policy of DBLP. Specifically, we focus on DBLP's ability to distinguish critical from non-critical learning phases, adapt loss tolerance accordingly, and maintain robust communication performance under transient microbursts. A three-worker setup is sufficient to exercise and observe this protocol behavior in a controlled manner. Before training begins, the server spawns three threads, each dedicated to a worker, accepts their incoming connections, and allocates unique UDP port numbers for communication. Evaluating DBLP at larger scales is left for future work.

At each training step, the server performs the reduce operation by collecting gradients from all workers and computing their mean. At the start of each epoch, DBLP checks whether the gradient norm change triggers CLR detection. If the current gradient triggers the CLR detection mechanism, the server adjusts the bounded-loss-tolerance threshold to $P_{low}$; otherwise, it sets it to $P_{high}$. The tolerance threshold is initialized to $P_{low}$, as the start of training is indisputably critical~\cite{achille2019criticallearningperiodsdeep}. Finally, the server broadcasts averaged gradients to all workers for model updates.

The pseudocode for the server and the worker is presented in Algorithms~\ref{alg:server} and~\ref{alg:worker} in the Appendix.

\begin{figure}
\centering
\begin{subfigure}{0.8\columnwidth}
    \centering
    \includegraphics[width=\linewidth]{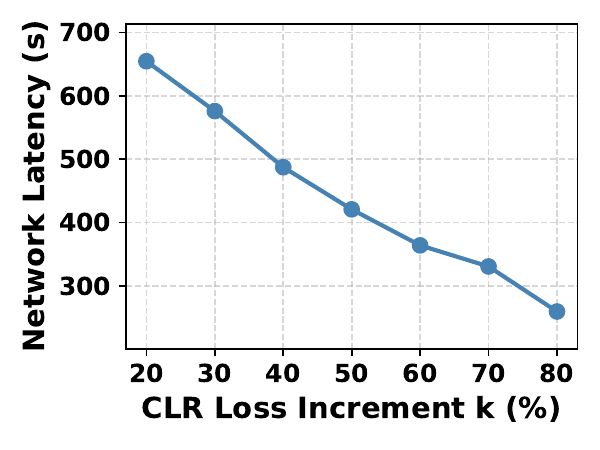}
    \caption{$P_{high}$ Sensitivity: Network Latency}
    \label{fig:phigh-latency}
\end{subfigure}
\begin{subfigure}{0.8\columnwidth}
    \centering
    \includegraphics[width=\linewidth]{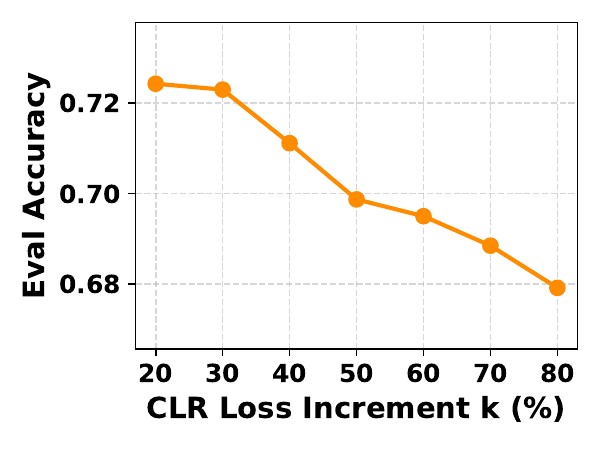}
    \caption{$P_{high}$ Sensitivity: Evaluation Accuracy}
    \label{fig:phigh-eval-acc}
\end{subfigure}
\caption{Sensitivity of DBLP to the non-CLR tolerance offset $k$ on EfficientNetB0, where $P_{high} = P_{low} + k$, sweeping $k \in \{20\%, 30\%, 40\%, 50\%, 60\%, 70\%, 80\%\}$. (a) Total send latency decreases smoothly and near-monotonically as $k$ grows. (b) Evaluation accuracy is stable up to $k = 30\%$ and then degrades. We adopt $k = 40\%$ as the default operating point: it captures substantial latency reduction while keeping evaluation accuracy above 71\%.}
\end{figure}

\section{Evaluation}

\subsection{Experimental Setup}
\subsubsection{Testbed}

We evaluate DBLP on a testbed consisting of 3 worker nodes and 1 server node, following the centralized all-reduce architecture. For image classification tasks, each worker node has 1 NVIDIA RTX A4500 GPU, 4 CPU cores, and 32 GB of memory. For language model benchmark, each worker node has 1 NVIDIA RTX 4090 GPU, 32 CPU cores, and 128 GB of memory. The network topology is treated as a black box to our system, and DBLP is designed to adapt to any settings.

\subsubsection{Models and Datasets}

EfficientNetB0~\cite{efficientnet} on CIFAR-10~\cite{cifar10}; ResNet50~\cite{resnet50} and AlexNet~\cite{alexnet} on CIFAR-100~\cite{cifar10}; GPT-2-S~\cite{GPT2} on WikiText-2~\cite{wikitext-2}.

As noted in Section~\ref{background-and-motivation}, modern LLM training involves complex communication patterns, including tensor parallelism~\cite{megatron-lm}, pipeline parallelism~\cite{pipedream}, expert parallelism~\cite{gshard}, and all-to-all collectives. Across all of these, gradient synchronization is a shared primitive that has been a recognized bottleneck since the early days of distributed DNN training~\cite{pipedream, bytescheduler}. DBLP addresses this primitive at the transport layer. Our experiments focus on CNNs and a small-scale language model due to hardware constraints, but the protocol behavior generalizes across model architectures.

\subsubsection{Baseline and Metrics}

We compare DBLP with a baseline across all experiments. For the baseline, we adopt a fixed bounded-loss tolerance $p$, following the suggested value from~\cite{mlt}, which was shown to maintain model performance. For DBLP, we set the loss tolerance to the same $p$ as the baseline within the critical learning regime (CLR). Outside the CLR, DBLP adaptively increases its tolerance to $p + k$, where the value of $ k $ is based on our empirical observations in Figure~\ref{fig:preliminary-densenet}. We choose $ k = 40\% $ as our preliminary experiments reveal that dropping $40\%$ gradients in non-critical phases does not noticeably affect convergence. We further justify our choice of $k$ in our sensitivity experiments in Section~\ref{eval: phigh-sensitivity}.

\subsubsection{\texorpdfstring{$P_{high}$}{P high} Sensitivity}
\label{eval: phigh-sensitivity}

We evaluate DBLP with $k$ values of $20\%$, $30\%$, $40\%$, $50\%$, $60\%$, $70\%$, and $80\%$ on EfficientNetB0. Setting the value of $k$ involves a trade-off between communication latency and model performance. Higher $k$ allows more aggressive gradient dropping, reducing latency at the cost of accuracy. Lower $k$ preserves more gradients at higher communication cost.

Figures~\ref{fig:phigh-latency} and~\ref{fig:phigh-eval-acc} show total network latency and final evaluation accuracy across the sweep. As expected, higher values of $k$ yield lower latency but also lower evaluation accuracy. The latency curve in Figure~\ref{fig:phigh-latency} decreases smoothly and near-monotonically with each increment of $k$. The accuracy curve in Figure~\ref{fig:phigh-eval-acc} remains flat between $20\%$ and $30\%$ (within $0.001$) and then degrades from $40\%$ to $80\%$. We choose a relatively conservative $k = 40\%$ as the preferred operating point because it captures a substantial fraction of the available latency reduction (approximately $25\%$ relative to $k = 20\%$) while keeping evaluation accuracy above $71\%$. 

Setting $k$ beyond $40\%$ trades accuracy at a markedly worse rate per unit of latency saved, whereas setting $k$ below $40\%$ forgoes meaningful latency gains without a corresponding improvement in accuracy. We therefore adopt $k = 40\%$ as the default in our remaining experiments. Tolerance values for all models are listed in Table~\ref{table:mlt-dblp-tolerance}. 

\begin{table*}
\centering
\begin{minipage}{0.48\linewidth}
  \centering
  \setlength{\tabcolsep}{6pt}
  \renewcommand{\arraystretch}{1.2}
  \begin{tabular}{lccc}
  \toprule
  \textbf{Latency} & \textbf{DBLP} & \textbf{Baseline} & \textbf{Speedup} \\
  \midrule
  Microburst 1 (Iter. 372) & 0.4630 & 1.8506 & 4.00$\times$ \\
  Microburst 2 (Iter. 837) & 0.3996 & 2.0099 & 5.03$\times$ \\
  Tail                     & 1.4633 & 2.0099 & 1.37$\times$ \\
  Average                  & 0.6749 & 1.0389 & 1.54$\times$ \\
  \bottomrule
  \end{tabular}
  \caption{Per-iteration send latency on EfficientNetB0 under a 60\% microburst loss rate. DBLP delivers $\sim$4--5$\times$ speedups at burst iterations and $\sim$1.4--1.5$\times$ improvements on tail and average latency.}
  \label{tab:latency_comparison_effnet_60mb}
\end{minipage}
\hfill
\begin{minipage}{0.48\linewidth}
  \centering
  \setlength{\tabcolsep}{6pt}
  \renewcommand{\arraystretch}{1.2}
  \begin{tabular}{lccc}
  \toprule
  \textbf{Latency} & \textbf{DBLP} & \textbf{Baseline} & \textbf{Speedup} \\
  \midrule
  Microburst 1 (Iter. 279) & 0.4621 & 2.1629 & 4.68$\times$ \\
  Microburst 2 (Iter. 651) & 0.3619 & 2.1293 & 5.88$\times$ \\
  Tail                     & 1.1305 & 2.1629 & 1.91$\times$ \\
  Average                  & 0.5907 & 0.9236 & 1.56$\times$ \\
  \bottomrule
  \end{tabular}
  \caption{Per-iteration send latency on EfficientNetB0 under a 70\% microburst loss rate. DBLP delivers $\sim$4.7--5.9$\times$ speedups at burst iterations and $\sim$1.6--1.9$\times$ improvements on average and tail latency.}
  \label{tab:latency_comparison_effnet}
\end{minipage}
\\[1.2em]
\begin{minipage}{0.48\linewidth}
  \centering
  \setlength{\tabcolsep}{6pt}
  \renewcommand{\arraystretch}{1.2}
  \begin{tabular}{lccc}
  \toprule
  \textbf{Latency} & \textbf{DBLP} & \textbf{Baseline} & \textbf{Speedup} \\
  \midrule
  Microburst 1 (Iter. 372) & 0.6214 & 3.2114 & 5.17$\times$ \\
  Microburst 2 (Iter. 837) & 0.6173 & 3.1336 & 5.08$\times$ \\
  Tail                     & 1.5106 & 3.2114 & 2.13$\times$ \\
  Average                  & 0.5797 & 0.9904 & 1.71$\times$ \\
  \bottomrule
  \end{tabular}
  \caption{Per-iteration send latency on EfficientNetB0 under an 80\% microburst loss rate. DBLP delivers $\sim$5$\times$ speedups at burst iterations, 2.13$\times$ lower tail latency, and 1.71$\times$ lower average latency.}
  \label{tab:latency_comparison_effnet_80mb}
\end{minipage}
\hfill
\begin{minipage}{0.48\linewidth}
  \centering
  \setlength{\tabcolsep}{6pt}
  \renewcommand{\arraystretch}{1.2}
  \begin{tabular}{lccc}
  \toprule
  \textbf{Latency} & \textbf{DBLP} & \textbf{Baseline} & \textbf{Speedup} \\
  \midrule
  Microburst 1 (Iter. 465) & 1.1686 & 6.5372 & 5.59$\times$ \\
  Microburst 2 (Iter. 558) & 1.1053 & 6.6004 & 5.97$\times$ \\
  Tail                     & 1.5824 & 6.6004 & 4.17$\times$ \\
  Average                  & 0.5792 & 1.1283 & 1.95$\times$ \\
  \bottomrule
  \end{tabular}
  \caption{Per-iteration send latency on EfficientNetB0 under a 90\% microburst loss rate, the most aggressive evaluated. DBLP achieves up to 5.97$\times$ speedups at burst iterations, 4.17$\times$ lower tail latency, and 1.95$\times$ lower average latency.}
  \label{tab:latency_comparison_effnet_90mb}
\end{minipage}
\end{table*}

\begin{table}[t]
\centering
\setlength{\tabcolsep}{6pt}
\renewcommand{\arraystretch}{1.2}
\begin{tabular}{lccc}
\toprule
\textbf{Latency} & \textbf{DBLP} & \textbf{Baseline} & \textbf{Speedup} \\
\midrule
Microburst 1 (Iter. 372)  & 2.3017 & 10.1180 & 4.40$\times$ \\
Microburst 2 (Iter. 651)  & 2.4367 & 10.0079 & 4.11$\times$ \\
Microburst 3 (Iter. 1209) & 2.6885 & 10.0118 & 3.72$\times$ \\
Tail                      & 5.6665 & 10.1180 & 1.79$\times$ \\
Average                   & 3.1966 & 4.8519  & 1.52$\times$ \\
\bottomrule
\end{tabular}
\caption{Per-iteration send latency on ResNet50 under a 70\% microburst loss rate (15 epochs, 1395 iterations). DBLP sustains 3.72--4.4$\times$ speedups at every burst iteration and $\sim$1.5--1.8$\times$ improvements on average and tail latency, demonstrating that burst resilience generalizes beyond EfficientNetB0.}
\label{tab:latency_comparison_resnet}
\end{table}

\begin{figure*}
\centering
\begin{subfigure}{0.49\textwidth}
    \centering
    \includegraphics[width=\linewidth]{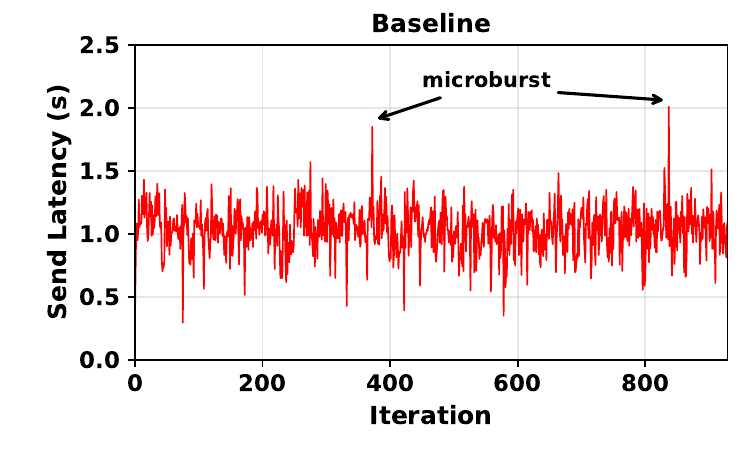}
    \caption{60\% microburst; fixed 0.8\% tolerance}
\end{subfigure}
\hfill
\begin{subfigure}{0.49\textwidth}
    \centering
    \includegraphics[width=\linewidth]{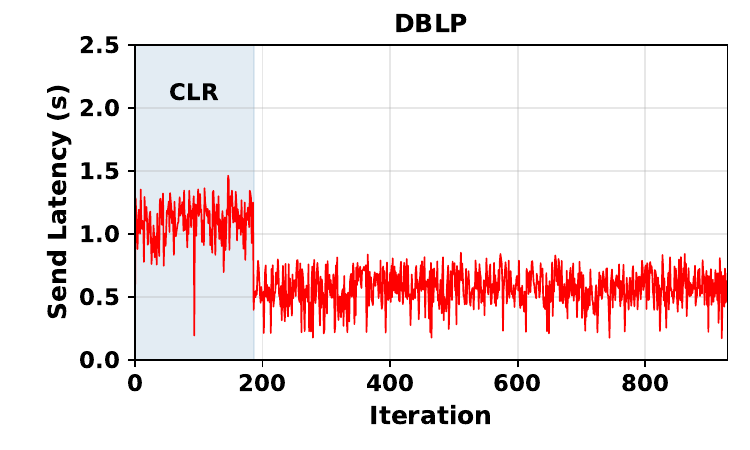}
    \caption{60\% microburst; CLR: 0.8\%, non-CLR: 40.8\%}
\end{subfigure}
\\[0.2em]
\begin{subfigure}{0.49\textwidth}
    \centering
    \includegraphics[width=\linewidth]{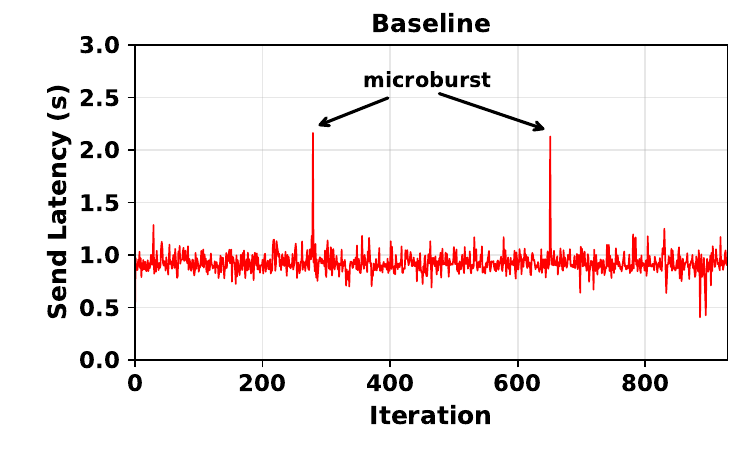}
    \caption{70\% microburst; fixed 0.8\% tolerance}
\end{subfigure}
\hfill
\begin{subfigure}{0.49\textwidth}
    \centering
    \includegraphics[width=\linewidth]{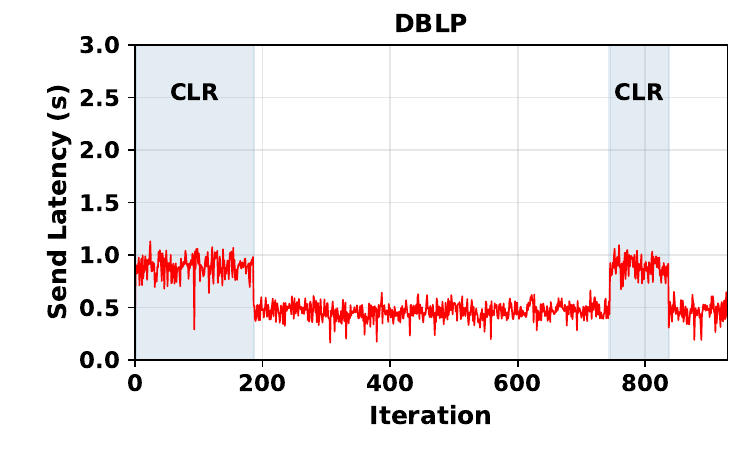}
    \caption{70\% microburst; CLR: 0.8\%, non-CLR: 40.8\%}
\end{subfigure}
\\[0.2em]
\begin{subfigure}{0.49\textwidth}
    \centering
    \includegraphics[width=\linewidth]{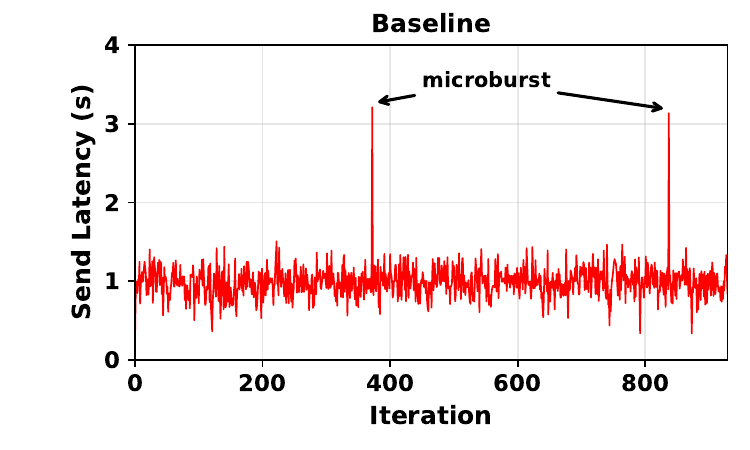}
    \caption{80\% microburst; fixed 0.8\% tolerance}
\end{subfigure}
\hfill
\begin{subfigure}{0.49\textwidth}
    \centering
    \includegraphics[width=\linewidth]{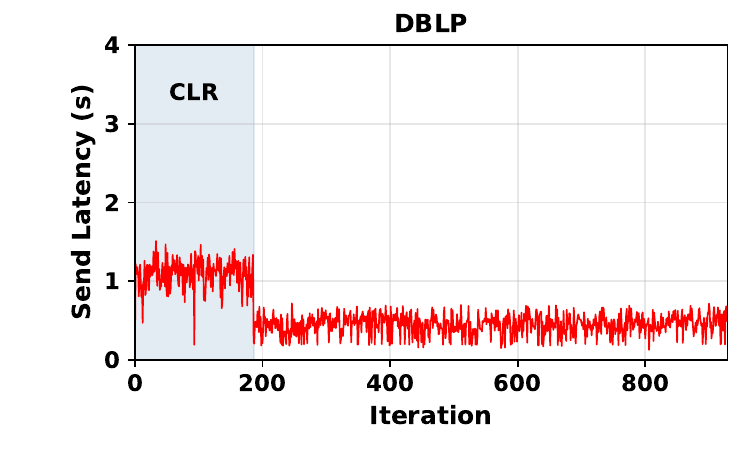}
    \caption{80\% microburst; CLR: 0.8\%, non-CLR: 40.8\%}
\end{subfigure}
\\[0.2em]
\begin{subfigure}{0.49\textwidth}
    \centering
    \includegraphics[width=\linewidth]{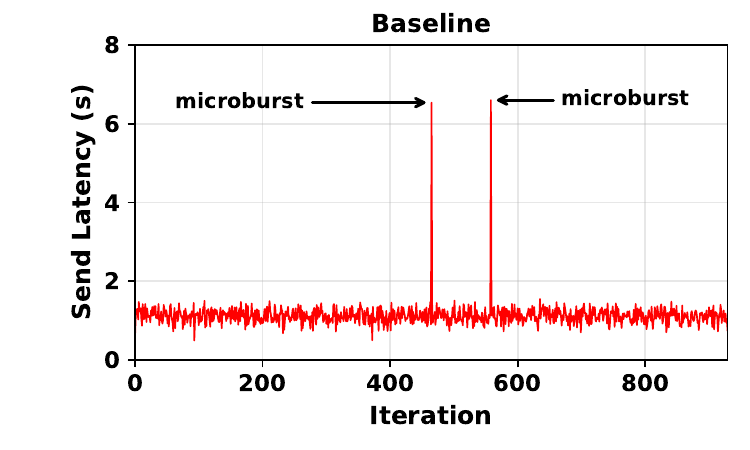}
    \caption{90\% microburst; fixed 0.8\% tolerance}
\end{subfigure}
\hfill
\begin{subfigure}{0.49\textwidth}
    \centering
    \includegraphics[width=\linewidth]{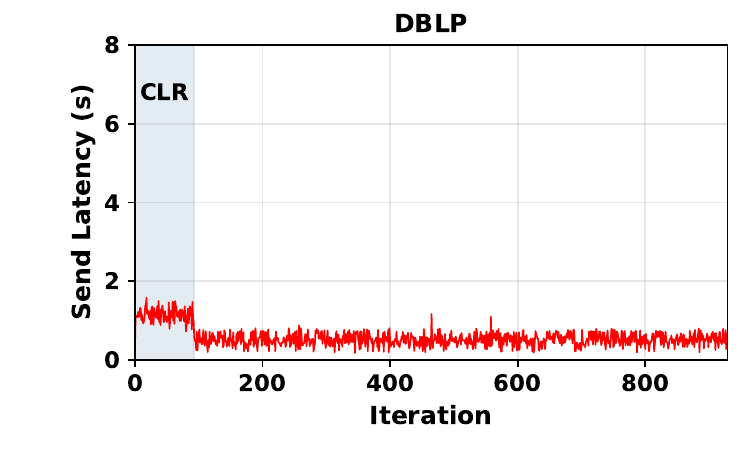}
    \caption{90\% microburst; CLR: 0.8\%, non-CLR: 40.8\%}
\end{subfigure}
\caption{Per-iteration send latency on EfficientNetB0 under injected microbursts at loss rates of 60\%, 70\%, 80\%, and 90\%. Each row pairs the fixed-tolerance baseline (left) with DBLP (right) at the same burst intensity. The baseline exhibits pronounced latency spikes at every injected microburst iteration that grow more severe with burst intensity, whereas DBLP's higher tolerance in non-critical regions absorbs the bursts and keeps per-round latency close to the no-burst case.}
\label{fig:microburst-effnet}
\end{figure*}

\begin{figure*}[!t]
\centering
\begin{subfigure}{0.32\textwidth}
    \centering
    \includegraphics[width=\linewidth]{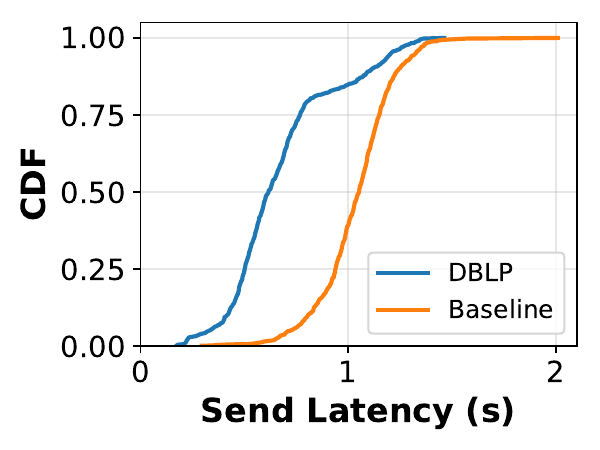}
    \caption{60\% MB, EfficientNetB0}
    \label{fig: 60-cdf-effent}
\end{subfigure}
\hfill
\begin{subfigure}{0.32\textwidth}
    \centering
    \includegraphics[width=\linewidth]{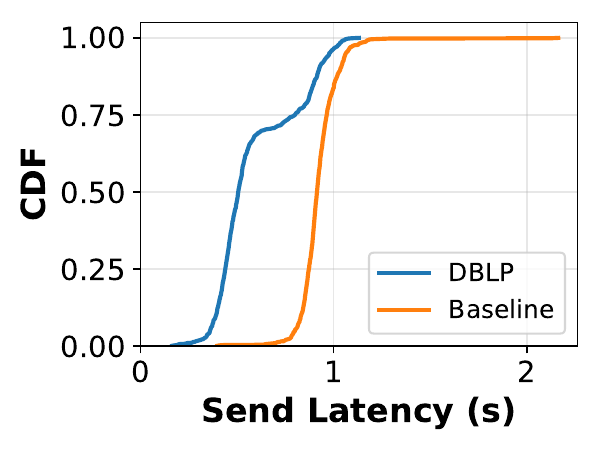}
    \caption{70\% MB, EfficientNetB0}
    \label{fig: 70-cdf-effent}
\end{subfigure}
\hfill
\begin{subfigure}{0.32\textwidth}
    \centering
    \includegraphics[width=\linewidth]{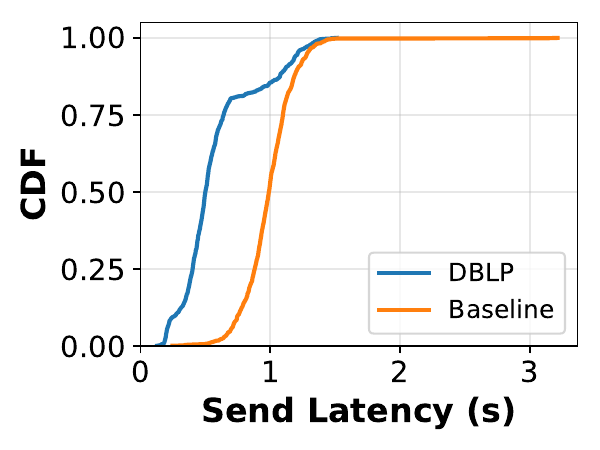}
    \caption{80\% MB, EfficientNetB0}
    \label{fig: 80-cdf-effent}
\end{subfigure}
\\[1ex]
\begin{subfigure}{0.32\textwidth}
    \centering
    \includegraphics[width=\linewidth]{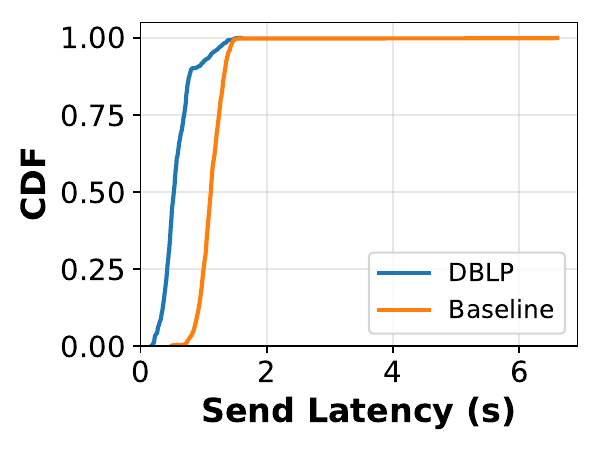}
    \caption{90\% MB, EfficientNetB0}
    \label{fig: 90-cdf-effent}
\end{subfigure}
\hspace{0.05\textwidth}
\begin{subfigure}{0.32\textwidth}
    \centering
    \includegraphics[width=\linewidth]{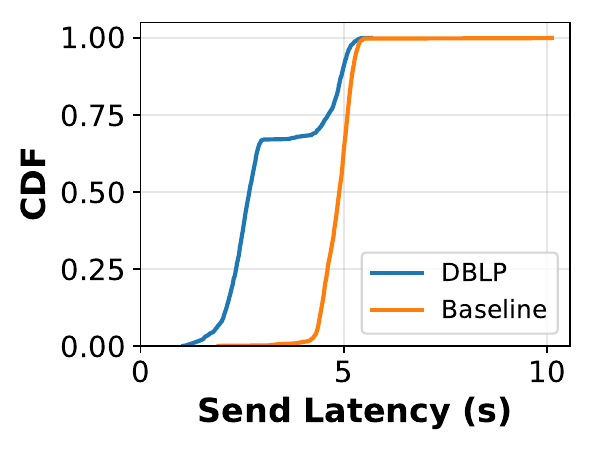}
    \caption{70\% MB, ResNet50}
    \label{fig:cdf-resnet}
\end{subfigure}
\caption{Cumulative distribution functions (CDFs) of per-iteration send latency under injected microbursts (MB). Panels (a)--(d) show EfficientNetB0 at 60\%, 70\%, 80\%, and 90\% microburst loss rates, and panel (e) shows ResNet50 at 70\%. Across every setting, DBLP's distribution sits to the left of the baseline and exhibits a substantially shorter right tail, indicating both lower average latency and dramatic tail latency reduction.}
\label{fig:cdf-effnet}
\end{figure*}

\begin{figure*}
\centering
\begin{subfigure}{0.49\textwidth}
    \centering
    \includegraphics[width=\linewidth]{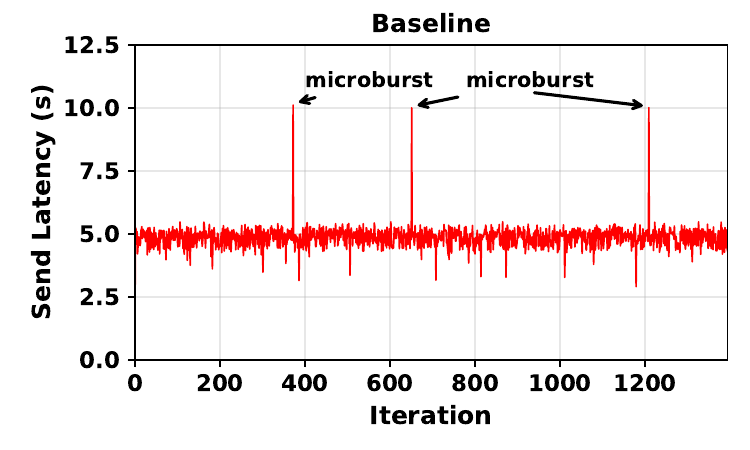}
    \caption{ResNet50: fixed 2.4\% tolerance}
\end{subfigure}
\hfill
\begin{subfigure}{0.49\textwidth}
    \centering
    \includegraphics[width=\linewidth]{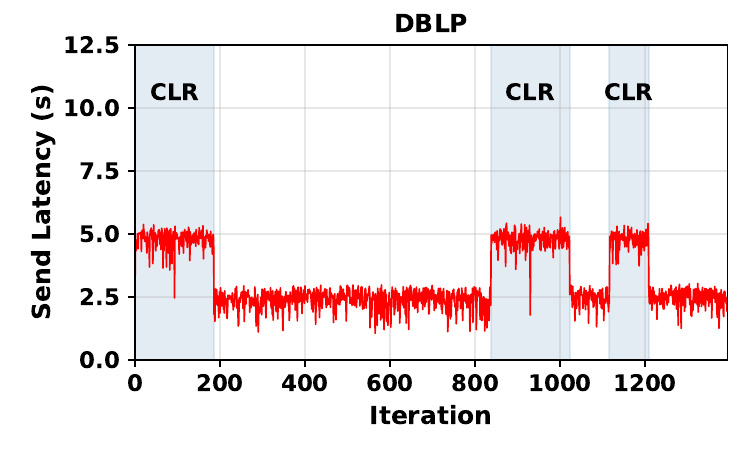}
    \caption{ResNet50: CLR: 2.4\%, non-CLR: 42.4\%}
\end{subfigure}
\vspace{0.5em}
\caption{Per-iteration send latency on ResNet50 under a 70\% microburst loss rate over 15 training epochs (1395 iterations). The baseline (left) exhibits sharp latency spikes at the injected microburst iterations 372, 651, and 1209, while DBLP (right) continues to deliver per-round latency comparable to the no-burst case, demonstrating that the protocol's burst resilience generalizes beyond EfficientNetB0.}
\label{fig:microburst-resnet}
\end{figure*}

\begin{table}[!htbp]
\centering

\setlength{\tabcolsep}{8pt}
\renewcommand{\arraystretch}{1.1}

\begin{tabular}{lcc}
\toprule
\textbf{Model} & \textbf{DBLP} & \textbf{Baseline} \\
\midrule
EfficientNetB0 (60\%) & $71.37\%$ & $71.77\%$ \\
EfficientNetB0 (70\%) & $70.78\%$ & $72.33\%$ \\
EfficientNetB0 (80\%) & $70.86\%$ & $72.21\%$ \\
EfficientNetB0 (90\%) & $70.41\%$ & $71.39\%$ \\
ResNet50 (70\%)       & $82.78\%$ & $83.54\%$ \\
\bottomrule
\end{tabular}

\caption{Final evaluation accuracy of DBLP and the fixed-tolerance baseline under injected microbursts: EfficientNetB0 at four loss rates (60--90\%) and ResNet50 at 70\%. Despite DBLP's substantially higher gradient loss in non-CLR phases, accuracy degradation is at most 1.55\% on EfficientNetB0 and 0.76\% on ResNet50, indicating that the large latency improvements come at minimal cost to model quality.}
\label{tab:microburst-eval-acc}
\end{table}

\begin{figure}
    \centering
    \includegraphics[width=\linewidth]{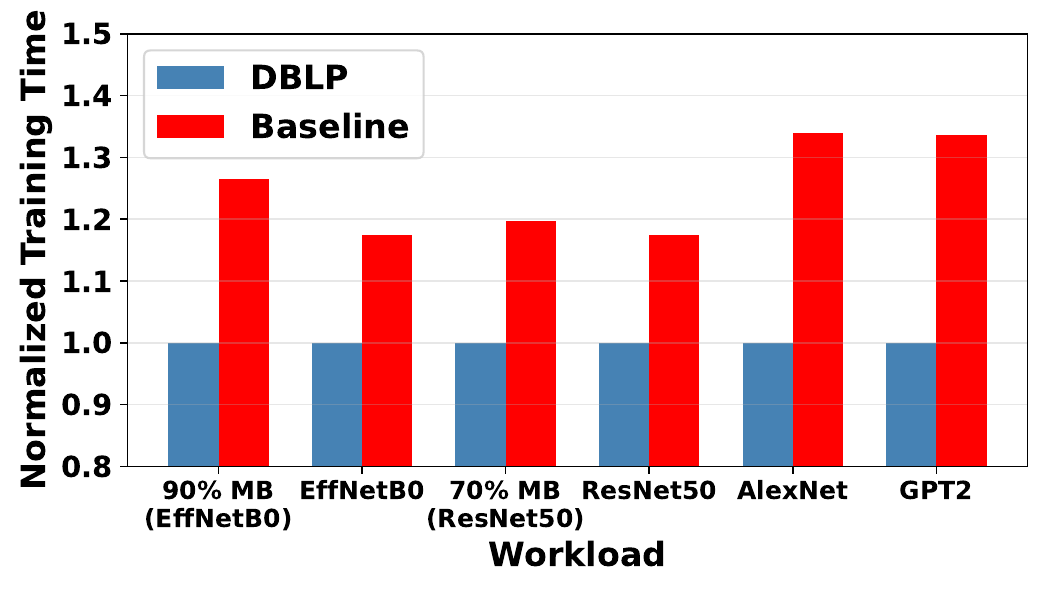}
    \caption{End-to-end training time of DBLP versus the fixed-tolerance baseline, normalized so that DBLP equals 1, across EfficientNetB0 (EffNetB0), ResNet50, AlexNet, and GPT-2-S, with and without injected microbursts (MB) where applicable. DBLP reduces total training time on every workload, by an average of 24.8\% and up to 33.9\%, with the largest gains observed on the larger AlexNet and GPT-2-S models.}
    \label{fig: training-time-normalized}
\end{figure}

\subsection{Microburst Results}
\label{eval: microburst}

The majority of congestion events in data centers are transient~\cite{micro-load-balancing, short-lived-1, short-lived-2}. Over $90\%$ of packet loss is caused by short-lived microbursts~\cite{micro-load-balancing, microburst}. Events where loss exceeds $50\%$ in a short period have also been observed~\cite{loss>50}. Therefore, we simulate microbursts by manually injecting bursty packet loss at rates $\in \{60, 70, 80, 90\}\%$, matching thresholds reported in prior work~\cite{loss>50}.

Specifically, we train EfficientNetB0 with all four loss rates for 10 epochs (i.e., 930 iterations). To demonstrate model-agnosticism, we also train ResNet50 at a $70\%$ loss rate for 15 epochs (i.e., 1395 iterations). We compare DBLP against the baseline: the baseline employs a constant loss tolerance, whereas DBLP adopts a $40\%$ higher tolerance in the non-critical learning region (non-CLR) for all microburst events. For each iteration, we record the send latency from workers to the server. We manually inject microbursts twice for EfficientNetB0 and three times for ResNet50; the iterations at which they occur are chosen at random and listed in Tables~\ref{tab:latency_comparison_effnet_60mb}--\ref{tab:latency_comparison_resnet}. These tables summarize the microburst (i.e., spike) latencies, tail latencies, and average latencies for DBLP and the baseline. We present the results for EfficientNetB0 in Figure~\ref{fig:microburst-effnet} and for ResNet50 in Figure~\ref{fig:microburst-resnet}. The CDF curves for EfficientNetB0 and ResNet50 are provided in Figure~\ref{fig:cdf-effnet}.

For EfficientNetB0, the baseline exhibits latency spikes at every injected microburst iteration regardless of the loss rate, whereas DBLP keeps per-round latency near the no-burst case with no visible spikes. Across the range of simulated loss rates, the latency collapse suffered by the baseline grows monotonically more severe as burst intensity increases. This is directly reflected in Figures~\ref{fig:microburst-effnet} and Tables~\ref{tab:latency_comparison_effnet_60mb}--\ref{tab:latency_comparison_effnet_90mb}: the send latency spikes of the baseline at the 90\% microburst iterations are approximately $3.57\times$ higher than those at the 60\% microburst iterations. Moreover, DBLP delivers $1.37\times$ lower tail latency at the $60\%$ loss rate and $4.17\times$ lower tail latency at the $90\%$ loss rate compared to the baseline. DBLP also achieves $4.00\times$ to $5.97\times$ faster communication, and $1.54\times$ to $1.95\times$ lower average latency than the baseline across loss rates from $60\%$ to $90\%$.

At higher loss rates, the baseline's fixed tolerance forces progressively more retransmission rounds, whereas DBLP's higher tolerance allows each communication round to finish early, preventing the buildup of long-tail latency. The $k = 40\%$ from Section~\ref{eval: phigh-sensitivity} is deliberately conservative and already proves robust enough to absorb microbursts. We hypothesize that the microburst improvements would be even more pronounced at values of $k > 40\%$, which we leave as a direction for future investigation.

For ResNet50, the latency spikes appear at the first iteration of epoch 4 (iter. 372), 7 (iter. 651), and 13 (iter. 1209) on the baseline. DBLP continues to demonstrate robustness to microbursts without exhibiting any latency spikes. Table~\ref{tab:latency_comparison_resnet} summarizes the statistics for ResNet50 using the same metrics, and a CDF curve is presented in Figure~\ref{fig:cdf-resnet}.

Despite the claim in~\cite{mlt} that their bounded-loss-tolerance gradient communication scheme resolves the long-tail latency issue, we find that in scenarios without switch-level support and under microbursts, a solution that deploys a fixed small tolerance value collapses. In contrast, DBLP, a fully software-based protocol, demonstrates latency resilience against bursty packet loss and accuracy robustness to high gradient loss tolerance, owing to its dynamic phase-adaptive property.

We present the evaluation accuracy results for this microburst experiment in Table~\ref{tab:microburst-eval-acc}. For EfficientNetB0, DBLP incurs accuracy degradations of $0.40\%$, $1.55\%$, $1.35\%$, and $0.98\%$ at the $60\%$, $70\%$, $80\%$, and $90\%$ microburst intensities, respectively. For ResNet50, DBLP incurs only a $0.76\%$ accuracy degradation. Since the majority of training happens outside the CLR, the probability that a microburst collides with DBLP's CLR is low.

\subsection{Training Time Speedups}
\label{eval: training-time}

We also train all models without manually injecting microbursts. The results, shown in Figure~\ref{fig: training-time-normalized} and Table~\ref{tab:training_time_comparison}, demonstrate that DBLP consistently outperforms the baseline across all models.

For EfficientNetB0, DBLP achieves a $26.42\%$ end-to-end training time speedup under the $90\%$ microbursts and a $17.44\%$ speedup without bursts. The two short-lived microbursts constitute only $0.215\%$ of total communication cycles, yet result in another $8.98\%$ latency degradation for the baseline. For ResNet50, DBLP achieves a $19.78\%$ reduction in training time under microbursts, and a $17.51\%$ reduction in the absence of bursts. DBLP also achieves end-to-end training time speedups of $33.9\%$ on AlexNet, and $33.6\%$ on GPT-2-S.

\begin{table}[!htbp]
\centering

\setlength{\tabcolsep}{8pt}
\renewcommand{\arraystretch}{1.1}

\begin{tabular}{lcc}
\toprule
\textbf{Experiment Type} & \textbf{DBLP} & \textbf{Baseline} \\
\midrule
90\% Microburst (EfficientNetB0) & 1.00 & 1.2642 \\
EfficientNetB0              & 1.00 & 1.1744 \\
70\% Microburst (ResNet50)       & 1.00 & 1.1978 \\
ResNet50                    & 1.00 & 1.1751 \\
AlexNet                     & 1.00 & 1.3393 \\
GPT-2                       & 1.00 & 1.3357 \\
\bottomrule
\end{tabular}

\caption{End-to-end training time of DBLP and the fixed-tolerance baseline, normalized so that DBLP equals 1, across two microburst-injected runs (90\% MB on EfficientNetB0, 70\% MB on ResNet50) and four clean runs (EfficientNetB0, ResNet50, AlexNet, GPT-2). DBLP reduces total training time on every workload by 17--34\%, with the largest gains on AlexNet (33.9\%) and GPT-2 (33.6\%).}
\label{tab:training_time_comparison}
\end{table}

\subsection{Evaluation Accuracy Results}
\label{eval: eval-acc}

\begin{figure*}
\centering
\begin{subfigure}{0.24\textwidth}
    \centering
    \includegraphics[width=\linewidth]{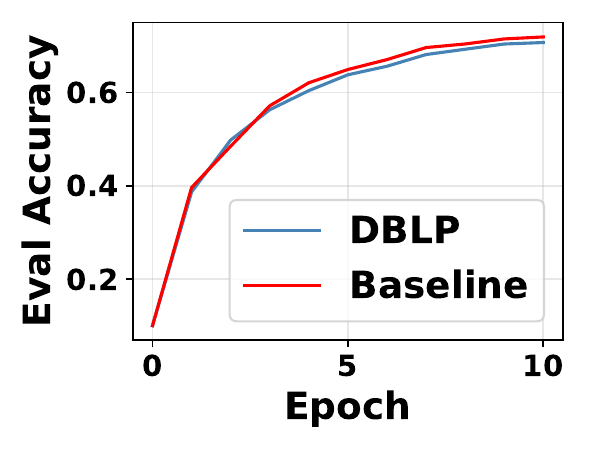}
    \caption{EfficientNetB0}
\end{subfigure}
\hfill
\begin{subfigure}{0.24\textwidth}
    \centering
    \includegraphics[width=\linewidth]{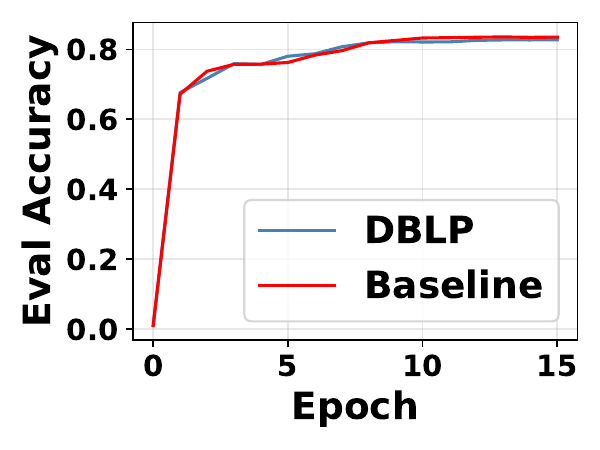}
    \caption{ResNet50}
\end{subfigure}
\hfill
\begin{subfigure}{0.24\textwidth}
    \centering
    \includegraphics[width=\linewidth]{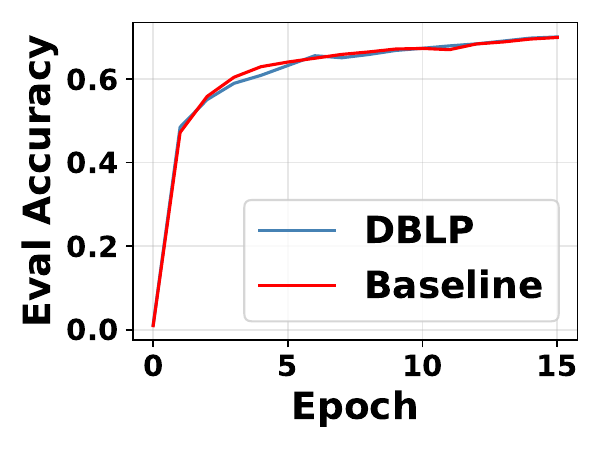}
    \caption{AlexNet}
\end{subfigure}
\hfill
\begin{subfigure}{0.24\textwidth}
    \centering
    \includegraphics[width=\linewidth]{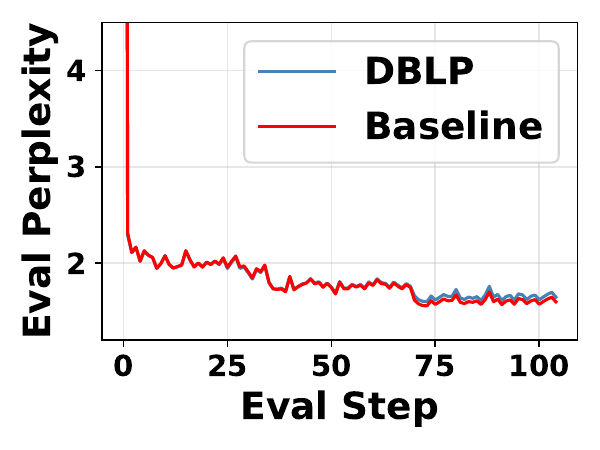}
    \caption{GPT-2-S}
    \label{fig:eval-acc-gpt}
\end{subfigure}
\caption{Evaluation curves of DBLP versus the fixed-tolerance baseline across four models trained without injected microbursts: (a) EfficientNetB0 on CIFAR-10, (b) ResNet50 on CIFAR-100, (c) AlexNet on CIFAR-100, and (d) GPT-2-S on WikiText-2 (perplexity, lower is better). Despite tolerating 40\% more gradient loss than the baseline in non-CLR phases, DBLP tracks the baseline closely on every model and converges to within a fraction of a percentage point of the baseline by the end of training.}
\label{fig:eta}
\end{figure*}

\begin{table}[!htbp]
\centering

\setlength{\tabcolsep}{10pt}
\renewcommand{\arraystretch}{1.1}

\begin{tabular}{lcc}
\toprule
\textbf{Model} & \textbf{DBLP} & \textbf{Baseline} \\
\midrule
EfficientNetB0 & 71.01\% & 71.30\% \\
ResNet50       & 82.74\% & 83.22\% \\
AlexNet        & 70.10\% & 69.97\% \\
GPT-2-S        & 1.6463  & 1.5981  \\
\bottomrule
\end{tabular}

\caption{Final test accuracy (EfficientNetB0, ResNet50, AlexNet) or perplexity (GPT-2-S, where lower is better) of DBLP versus the fixed-tolerance baseline, all trained without injected microbursts. DBLP achieves at least 99.4\% of the baseline's accuracy on the CNN models and comparable perplexity on GPT-2-S, confirming that the 40\% additional gradient loss tolerated in non-CLR phases does not meaningfully impair convergence.}
\label{tab:accuracy_comparison}
\end{table}

Figure~\ref{fig:eta} shows evaluation accuracy of DBLP versus the baseline. Despite a $40\%$ higher gradient loss rate in non-critical phases, DBLP maintains nearly the same accuracy as the baseline across all models. For EfficientNetB0, DBLP achieves $99.59\%$ of the baseline accuracy. For ResNet50, DBLP achieves $99.42\%$. Both DBLP and the baseline converge to approximately the same accuracy on AlexNet.

For GPT-2-S, we use perplexity as the evaluation metric, where lower values indicate better performance. Both converge to approximately the same perplexity, as shown in Figure~\ref{fig:eval-acc-gpt}. The test accuracy/perplexity values for all models are shown in Table~\ref{tab:accuracy_comparison}.

It is worth noting that microbursts do not necessarily cause the baseline to converge to lower accuracy: the total percentage of gradients received is unchanged, but the time to deliver all the gradient packets increases dramatically.

While we observe pronounced tail latency spikes in the baseline that significantly increase its training time, the baseline still receives and processes $99.2\%$ of gradient data at every training iteration (because of the fixed $0.8\%$ tolerance threshold for EfficientNetB0 as an example). Besides, test accuracy converges within a stable range and may fluctuate within that range, which explains why some evaluation accuracies in the microburst experiments are higher than the no-burst results.

\section{Future Work}
\label{future-work}

\subsection{Bounded-loss Cross-DC Collectives}

Recent efforts in NCCL have introduced topology-aware optimizations for cross-datacenter training, reducing traffic over slower inter-DC links and improving communication efficiency~\cite{gillis2025nccl_cross_dc}. However, these systems still assume exact collective semantics and do not account for how a model’s sensitivity to communication errors can change during training. One direction is to combine topology awareness with training-aware transport, enabling DBLP to dynamically adjust reliability across datacenter boundaries through its bounded-loss and phase-adaptive mechanisms. This could help absorb the jitter, bursts, and synchronization stalls that emerge once collectives span inter-DC distances, while maintaining convergence.

\subsection{Network Scheduling}

We discover that model training is highly sensitive only during the critical learning regime (CLR). This insight suggests an opportunity for datacenter network scheduling: allocating higher priority and bandwidth to training flows within their CLRs. Together with training-aware transport, such policies could improve overall cluster efficiency by dynamically prioritizing the most convergence-critical flows.

\section{Conclusion}

As distributed training continues to scale (up within racks, out across clusters, and increasingly across sites), communication reliability can no longer be treated as a purely network-layer concern independent of model behavior. Our study shows that the interaction between training dynamics and transport decisions fundamentally shapes system performance under realistic datacenter conditions. In particular, transient congestion events and microbursts expose the limitations of static reliability policies that treat all gradients and all training phases uniformly.

DBLP demonstrates that communication protocols for distributed model training can be designed to be phase-aware rather than phase-agnostic. By coupling bounded-loss transmission with critical learning regime detection, DBLP adapts reliability in response to training sensitivity, allowing the system to relax retransmissions when safe and enforce stricter delivery when necessary. This alignment between learning dynamics and transport control enables substantial reductions in tail latency when microbursts occur and in training time while preserving model convergence. Our results suggest that incorporating training-phase awareness into transport-layer decisions offers a practical and scalable path toward improving large-scale ML systems.

\bibliographystyle{IEEEtran}
\bibliography{main_paper}

\appendix
\subsection{Algorithm Pseudocode}

\begin{algorithm}
\caption{DBLP Send}
\label{alg:dblp_send}
\small
\begin{algorithmic}[1]
\Require Data to send
\State Split data into chunks
\State Initialize $bitmap \gets 0$
\While{no stop signal}
  \For{each unsent chunk}
    \If{stop signal arrives}
      \State return
    \EndIf
    \State Send chunk
  \EndFor
  \State Send probe signal $P$ through TCP channel
  \If{upon receiving the bitmap signal (B)}
    \State $bitmap \gets$ bitmap data
  \ElsIf{upon receiving the stop signal (S)}
    \State return
  \EndIf
\EndWhile
\end{algorithmic}
\end{algorithm}

\begin{algorithm}
\caption{DBLP\_RECV}
\label{alg:dblp_recv}
\small
\begin{algorithmic}[1]
\Ensure Data received up to the current tolerance threshold
\State $bitmap \gets 0$
\State $stop \gets \textbf{false}$
\While{\textbf{not} $stop$}
  \If{a new chunk arrives via UDP channel}
    \State Perform chunk integrity check
    \If{the chunk is valid}
      \State Accept the chunk and update $bitmap$
    \EndIf
  \EndIf
  \If{Probe $P$ is received}
    \If{loss-tolerance threshold is satisfied}
        \State Send stop signal $S$
        \State $stop \gets \textbf{true}$
        \State continue
    \Else
        \State Send bitmap signal $B$ with $bitmap$
    \EndIf
  \Else
    \If{loss-tolerance threshold is satisfied}
        \State Send stop signal $S$
        \State $stop \gets \textbf{true}$
        \State continue
    \EndIf
  \EndIf
\EndWhile
\State Unreceived chunks remain 0-filled.
\State Reconstruct the original data.
\State \textbf{return} reconstructed data
\end{algorithmic}
\end{algorithm}

\begin{algorithm}
\caption{Server}
\label{alg:server}
\small
\begin{algorithmic}[1]
\Require Number of workers $N$
\Require $P_{high}$ and $P_{low}$ (CLR upper/lower tolerance)
\For{each worker $i \in \{1,\dots,N\}$}
  \State Spawn a thread for worker $i$
  \State Accept a TCP connection from worker $i$
  \State $step \gets 0$
  \While{the worker connection is alive}
    \State $\nabla \mathbf{w}_i \gets \textsc{DBLP\_RECV}()$
    \State Wait until all workers' gradients arrive
    \State $\widehat{\nabla \mathbf{w}} \gets \operatorname{mean}(\nabla \mathbf{w}_1,\nabla \mathbf{w}_2,\dots,\nabla \mathbf{w}_N)$
    \If{at the beginning of an epoch}
        \State Reset loss-tolerance threshold to $P_{high}$
        \If{CLR is detected}
        \State Set loss-tolerance threshold to $P_{low}$
        \EndIf
    \EndIf
    \State \textsc{DBLP\_SEND}($\widehat{\nabla \mathbf{w}}$)
    \State $step \gets step + 1$
  \EndWhile
\EndFor
\end{algorithmic}
\end{algorithm}

\begin{algorithm}
\caption{Worker}
\label{alg:worker}
\small
\begin{algorithmic}[1]
\Require Server TCP socket address
\State Connect to the server
\For{each epoch}
  \For{each mini-batch}
    \State $\nabla \mathbf{w} \gets$ forward and backward pass
    \State \textsc{DBLP\_SEND}($\nabla \mathbf{w}$)
    \State $\widehat{\nabla \mathbf{w}} \gets \textsc{DBLP\_RECV}()$
    \State $\mathbf{w} \gets \mathbf{w} - \alpha \widehat{\nabla \mathbf{w}}$
  \EndFor
\EndFor
\end{algorithmic}
\end{algorithm}

\end{document}